\RequirePackage{fix-cm}
\documentclass[smallextended]{svjour3}       % onecolumn (second format)
\smartqed  % flush right qed marks, e.g. at end of proof
\usepackage{graphicx}
\usepackage{amssymb}
\usepackage{mathdots}
\usepackage[classicReIm]{kpfonts}
\usepackage{array}
\usepackage[square,numbers]{natbib}
\usepackage{subfigure}
\usepackage{wrapfig}
\usepackage{wasysym}
\usepackage{enumitem}
\usepackage{adjustbox}
\usepackage{ragged2e}
\usepackage[svgnames,table]{xcolor}
\usepackage{tikz}
\usepackage{longtable}
\usepackage{changepage}
\usepackage{setspace}
\usepackage{hhline}
\usepackage{multicol}
\usepackage{tabto}
\usepackage{float}
\usepackage{multirow}
\usepackage{makecell}
\usepackage{fancyhdr}
\usepackage[toc,page]{appendix}
\usepackage[hidelinks]{hyperref}
\usetikzlibrary{shapes.symbols,shapes.geometric,shadows,arrows.meta}
\tikzset{>={Latex[width=1.5mm,length=2mm]}}
\usepackage[utf8]{inputenc}
\usepackage[T1]{fontenc}
\TabPositions{0.14in,0.28in,0.42in,0.56in,0.7in,0.84in,0.98in,1.12in,1.26in,1.4in,1.54in,1.68in,1.82in,1.96in,2.1in,2.24in,2.38in,2.52in,2.66in,2.8in,2.94in,3.08in,3.22in,3.36in,3.5in,3.64in,3.78in,3.92in,4.06in,4.2in,4.34in,4.48in,4.62in,4.76in,4.9in,5.04in,5.18in,5.32in,5.46in,5.6in,5.74in,5.88in,6.02in,6.16in,6.3in,6.44in,6.58in,6.72in,6.86in,7.0in,7.14in,}
\begin{document}
%{\noindent\onecolumn{
%\noindent\textcopyright 2020 Springer. Personal use of this material is %permitted. Permission from Springer must be obtained for all other uses, %in any current or future media, including reprinting/republishing this %material for advertising or promotional purposes, creating new collective %works, for resale or redistribution to servers or lists, or reuse of any %copyrighted component of this work in other works.}}

%\newpage
\title{Beyond Background-Aware Correlation Filters: Adaptive Context Modeling by Hand-Crafted and Deep RGB Features for Visual Tracking}
%%%%%%%%%%%%%%%%%%%%%%%%%%%%%%%%%%%%%%%%%%%%%%%%%%%%%%%%%%%%%%%%%%%%%%%%%%%%%%%%%%%%%%%%%%%%%%%%%%%%%%%%%%
\author{Seyed~Mojtaba~Marvasti-Zadeh    \and
        Hossein~Ghanei-Yakhdan    \and
        Shohreh~Kasaei    
}
\institute{  \textbullet\ S. M. Marvasti-Zadeh \at
              Digital Image and Video Processing Lab (DIVPL), Department of Electrical Engineering, Yazd University, Yazd, Iran. He is also a member of Image Processing Lab (IPL), Sharif University of Technology, Tehran, Iran, and Vision and Learning Lab, University of Alberta, Edmonton, Canada (\email{mojtaba.marvasti@ualberta.ca}). \\  
             \textbullet\ H. Ghanei-Yakhdan (corresponding author) \at
              Digital Image and Video Processing Lab (DIVPL), Department of Electrical Engineering, Yazd University, Yazd, Iran (\email{hghaneiy@yazd.ac.ir}). \\
            \textbullet\ S. Kasaei \at
              Image Processing Lab (IPL), Department of Computer Engineering, Sharif University of Technology, Tehran, Iran (\email{kasaei@sharif.edu}).
}
%%%%%%%%%%%%%%%%%%%%%%%%%%%%%%%%%%%%%%%%%%%%%%%%%%%%%%%%%%%%%%%%%%%%%%%%%%%%%%%%%%%%%%%%%%%%%%%%%%%%%%%%%%
\date{Received: date / Accepted: date}
\maketitle
%%%%%%%%%%%%%%%%%%%%%%%%%%%%%%%%%%%%%%%%%%%%%%%%%%%%%%%%%%%%%%%%%%%%%%%%%%%%%%%%%%%%%%%%%%%%%%%%%%%%%%%%%%
\begin{abstract}
In recent years, the background-aware correlation filters have achie-ved a lot of research interest in the visual target tracking. However, these methods cannot suitably model the target appearance due to the exploitation of hand-crafted features. On the other hand, the recent deep learning-based visual tracking methods have provided a competitive performance along with extensive computations. In this paper, an adaptive background-aware correlation filter-based tracker is proposed that effectively models the target appearance by using either the histogram of oriented gradients (HOG) or convolutional neural network (CNN) feature maps. The proposed method exploits the fast 2D non-maximum suppression (NMS) algorithm and the semantic information comparison to detect challenging situations. When the HOG-based response map is not reliable, or the context region has a low semantic similarity with prior regions, the proposed method constructs the CNN context model to improve the target region estimation. Furthermore, the rejection option allows the proposed method to update the CNN context model only on valid regions. Comprehensive experimental results demonstrate that the proposed adaptive method clearly outperforms the accuracy and robustness of visual target tracking compared to the state-of-the-art methods on the OTB-50, OTB-100, TC-128, UAV-123, and VOT-2015 datasets. 
\end{abstract}

\section{Introduction}
\label{sec:1_Intro}
Visual object tracking is one of the challenging practical tasks in computer vision that has widespread applications in self-driving cars, autonomous robots, augmented reality, surveillance, and so forth. Although these methods have had significant progress in the constraint environments or with prior knowledge about the target, they are still faced with many challenging factors in practical scenarios; including occlusion, deformation, background clutter, camera motion, and out-of-view target, to name a few. One of the most critical issues of recent visual tracking methods is how to model the target appearance in order to be robust against challenging factors in real scenes. Hence, deep learning-based visual trackers are used that mostly exploit RGB feature maps extracted CNNs. Although the deep learning-based visual trackers obtain satisfactory results, they mostly suffer from a poor real-time tracking performance because of their expensive computations on all video frames \cite{MDNet,CCOT}. 

Generally speaking, the visual target trackers can be mainly categorized into discriminative \cite{VisComp_ImpCF,VisComp_NewTLDCF,VisComp_LongSTC,VisComp_PartSRDCF}, generative \cite{Generative1,Generative2}, and hybrid discriminative-generative \cite{STC,ASTCT} methods; based on their target appearance modeling. While the generative trackers learn a target model and search the best matching region in the subsequent frame, the discriminative trackers try to discriminate the target from its background. With the development of powerful discriminative methods in object detection, tracking-by-detection methods have achieved great success in visual target tracking. Discriminative correlation filters (DCF) are the most popular branch of tracking-by-detection that learn the target model via learning correlation filters by a set of training samples. Due to simultaneous visual tracking and filter training via the fast Fourier transform (FFT), correlation filter-based trackers are efficient in terms of both computation and memory.

On the other hand, the hybrid discriminative-generative trackers use the combination of generative and discriminative trackers to efficiently benefit from the background information. Background-aware (or context-aware) visual trackers have been widely used to improve the robustness of the target model. In general, these trackers either use the iterative optimization methods or formulate a closed-form solution to learn a context model. However, these trackers either prevent to use deep RGB features to preserve the computational efficiency of DCF or fuse the hand-crafted and deep RGB features which dramatically degrades tracking speed. Moreover, these trackers construct a context model based on all estimated target regions which their information validity has not been trusted. 

Motivated by the above problems, an adaptive background-aware DCF-based method is proposed which provides both the validation process of target representations and the adaptive learning procedure of two context models according to the difficulty of target estimation. To the best of our knowledge, this paper presents the first background-aware visual tracker that exploits the hand-crafted and deep RGB features, adaptively. The main contributions of the proposed method are summarized as follows.

1) Based on the difficulty of target state estimation, the proposed method exploits two context models; where one has low dimensional hand-crafted features (i.e., HOG context model), and the other includes robust deep RGB features with high computational complexity (i.e., CNN context model). Adaptive utilization of these context models helps the proposed tracker to not only exploit the CNN context model in challenging situations but also to have an acceptable speed compared to deep learning-based visual trackers. Furthermore, the simultaneous exploitation of low-level and high-level information improves the accuracy and robustness of the proposed method in realistic scenarios.

2) To detect the challenging situations that the HOG context model may not be accurately able to estimate target region, the proposed method utilizes the fast 2D-NMS algorithm \cite{2DNMS} and rejection option which sets by the semantic similarity comparison of target representations. 

3) With the aid of semantic similarity comparisons, the proposed method selects the best target estimation and updates the greedy CNN context model to localize the target when the context model with hand-crafted features is ineffective.

Comprehensive experimental results on five well-known visual tracking datasets demonstrate the effectiveness of the proposed method compared to the state-of-the-art visual tracking methods.

The rest of this paper is organized as follows. The related work is briefly explored in Section \ref{sec:2_RelatedWork}. The proposed adaptive background-aware visual tracking method is explained in Section \ref{sec:3_ProposedMethod}. Section \ref{sec:4_ExpResults} represents the experimental results. Finally, the conclusion is summarized in Section \ref{sec:5_Conclusion}.

\section{Related Work}
\label{sec:2_RelatedWork}
In this section, two types of the most well-known visual tracking methods including the correlation filters-based and context-aware visual trackers are briefly explained. The minimum output sum of squared error (MOSSE) correlation filters \cite{MOSSE} has engrossed the attention of researchers in the field of detection and tracking. The MOSSE is a simple visual tracking strategy that circumvents conventional problems with correlation filters and prevents the visual tracker to respond incorrectly to the background information. With the aid of efficient training of multi-channel detector/filter, the multi-channel correlation filters \cite{MultiChannelCF} extended the correlation filter theory to produce a single channel response map for a variety of tasks; such as detection and localization. Since then, this extension has been broadly used in other correlation filter-based visual trackers. To avoid iterating shifts of target patches to form a training set, the tracking-by-detection with Kernels method \cite{KCF-ECCV} exploited the well-established theory of Circulant matrices to derive closed-form training and detection solutions for different Kernels. Using the discrete Fourier transform (DFT), the Kernelized correlation filter (KCF) tracker \cite{KCF} developed an analytic model that can diagonalize a data matrix with thousands of translated target patches for high-speed tracking purposes.

Although correlation filter-based trackers have a significant computational efficiency, their detection/tracking performance suffers from unwanted boundary effects caused by the periodic assumption. Hence, correlation filters with limited boundaries \cite{CFLimitedBoundaries} utilized an augmented objective function which can largely remove the boundary effects. With the aid of the alternating direction method of multipliers (ADMMs) \cite{ADMM}, this tracker efficiently optimized the ridge regression objective function, iteratively. With a simple and powerful convex optimization strategy, the ADMM decomposed the augmented Lagrangian function to extremely efficient sub-problems that are appropriate for large learning problems. Also, the spatially regularized discriminative correlation filters (SRDCF) tracker \cite{SRDCF} formulated the correlation filters with a spatial regularization component that allows larger image regions to contribute to the online learning process.

On the other hand, the context-aware visual tracking method \cite{ContextAwareVisualTracking} utilized the context information via three main components; including auxiliary objects, a collaborative framework, and robust fusion. The context tracker \cite{ContextTracker} automatically explored supporters and distracters on-the-fly to verify the genuine target and avoid drifting. The spatio-temporal structural context-based tracker (STT) \cite{Wen2012} utilized the temporal and spatial context models that were constructed by the linear subspace method and the contributors (like supporters). To improve the robustness of target appearance variations, the fast spatio-temporal context tracking method (STC) \cite{STC} constructed spatial and spatio-temporal context models with fast learning and detection in the frequency domain. To track generic targets robustly in challenging situations, the adaptive STC method \cite{ASTCT} utilized adaptive learning parameters determination, accurate target localization with alternating feature set, and modified scale estimation scheme. Also, the context-aware correlation filter tracking \cite{ContextAwareCFTracking} provides a closed-form solution to boost the performance of correlation filter-based trackers by using the context information. Using densely extracted real negative training samples (instead of shifted foreground patches), the background-aware correlation filters (BACF) \cite{BACF} learn optimized correlation filters by using the ADMM and then update the context model with Sherman-Morrison lemma to enhance the accuracy and robustness of visual tracking.

Recently, some correlation filter-based trackers have been proposed to exploit deep CNN features. With the aid of multiple levels of CNN features, the hierarchical correlation features-based tracker (HCFT) \cite{HCFT} uses a coarse-to-fine fashion on achieved CNN response maps to handle challenging scenarios of visual tracking. In addition to the exploitation of hierarchical CNN features, the advanced version of HCFT (i.e., HCFT* of HCFTs) \cite{HCFTs} uses region proposals and a discriminative classifier to handle tracking failures (or the drift problem). The long-term correlation tracking (LCTdeep) \cite{LCTdeep} uses a passive-aggressive algorithm, multiple adaptive correlation filters, and incorporation of hand-crafted and deep CNN features for robust visual tracking. Inspired from \cite{SiamFC}, which uses fully convolutional Siamese networks for visual tracking, the discriminative correlation filter-based tracker (DCFNet) \cite{DCFNet} trains a Siamese network with a correlation layer to not only utilize the prior knowledge of convolutional layers and online feature learning, but also to benefit from the efficiency property of correlation filters. 

Although these visual tracking methods have provided competitive results for generic object tracking, they still suffer from some drawbacks. The conventional context-aware correlation filters usually exploit hand-crafted features (which might have a lower representational power compared to the deep CNN features) to preserve the computation efficiency of correlation filter-based methods. Besides, other methods that exploit deep CNN features either alternate the architecture of deep neural networks to provide computational efficiency or combine different features with improving the target model representation in all video frames. Despite the existing methods, the proposed method adaptively exploits hand-crafted and deep RGB features in two independent target models. With the aid of two context models, context validation, and decision making based on the comparison of semantic similarities, the proposed method provides a superior visual tracking performance with an efficient computational cost.

\section{Proposed Method}
\label{sec:3_ProposedMethod}
\begin{figure}
\includegraphics[width=11.9cm, height=7.5cm]{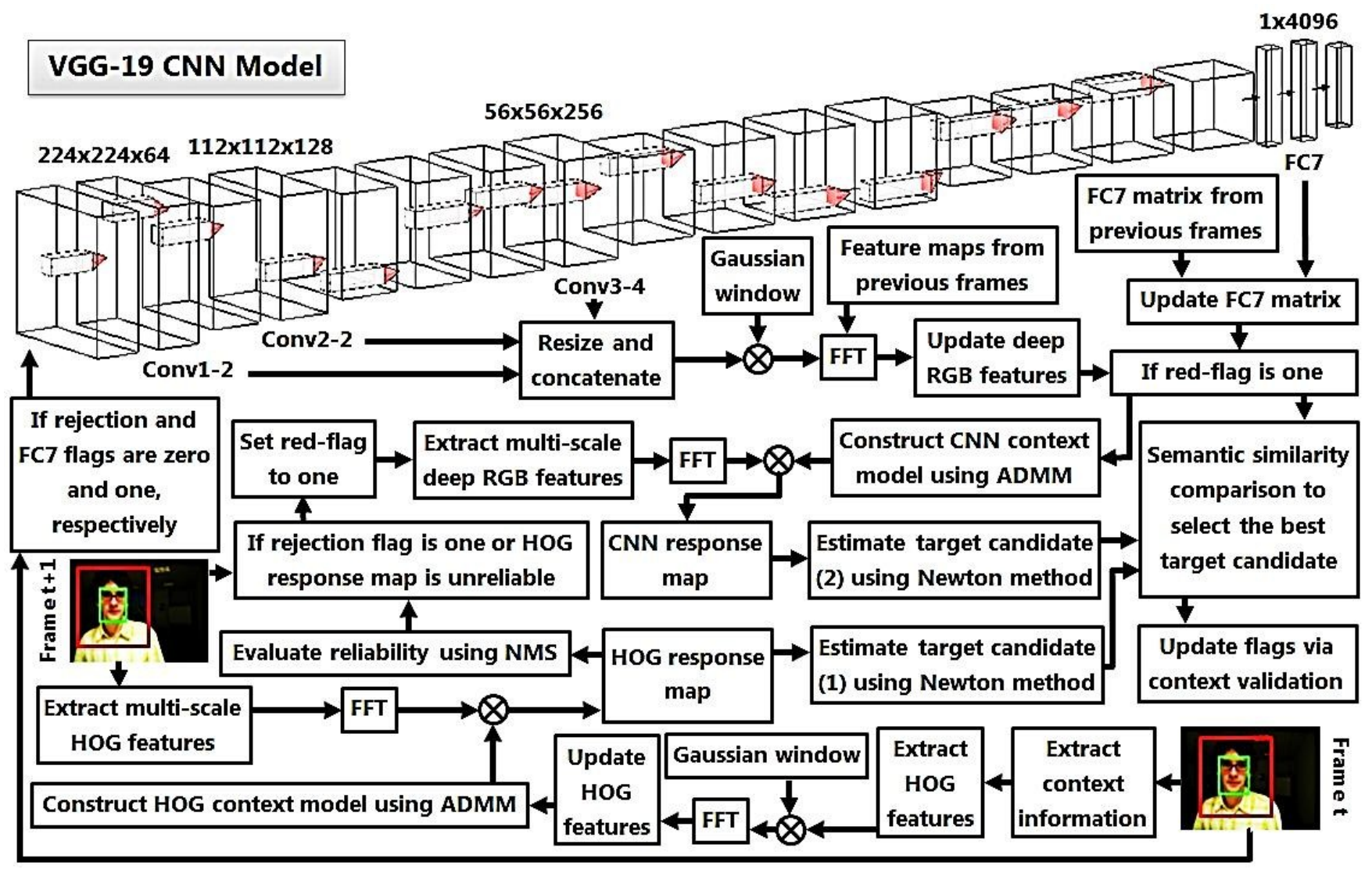}
%\vspace{-1mm}
\caption{Overview of proposed adaptive background-aware visual tracking method.}\label{Fig1_Overview}
\vspace{-4mm}
\end{figure}

In this section, the proposed adaptive BACF filters for visual tracking are presented that improve the BACF tracking performance, effectively. The proposed method is roughly decomposed to three main components. First, the adaptive context modeling is explained. Next, the validation process of context information is introduced. Finally, decision making and update procedure are described. The overview of the proposed method is shown in Fig. \ref{Fig1_Overview}. 

\subsection{Adaptive Context Modeling}
\label{sec:3_1}
Although BACF improves the tracking robustness against challenging attributes (e.g., deformation, scaling, and background clutter) with the incorporation of background information as the real negative examples, this method might quickly fail when their response maps do not generate a strict peak for the target location. This problem mostly comes from the exclusive exploitation of hand-crafted features which provide limited representation power for target modeling. Hence, the proposed method exploits the HOG and CNN context models, adaptively. Although the BACF tracker only learns the HOG context model, the proposed method exploits the HOG or Deep RGB features to learn two context models, independently. 

The two multi-channel context models are trained by minimizing the following objective function \cite{BACF} in the frequency domain

\begin{equation}\label{Eq.1}
\begin{array}{c}
{\mathop{\mathrm{min}}_{\mathrm{w,}\hat{z}}\mathrm{\ \ } \frac{\mathrm{1}}{2}{\left\|\hat{y}-\hat{X}\hat{z}\right\|}^2_2+\frac{\lambda }{2}{\left\|w\right\|}^2_2\ } \\ 
s.t.\ \ \ \ \hat{z}=\sqrt{M}\left(FB^T\otimes I_N\right)\ w \end{array}
\end{equation}
in which $\hat{}$ defines the DFT such that $F$ is the Fourier transform matrix (including orthonormal and complex basis vectors) for mapping to the frequency domain, and $T$ denotes the transpose operator on a complex matrix. Moreover, $\hat{y}$, $\hat{X}$, $\hat{z}$, $w$, $\lambda$ and $I_N$ denote the desired correlation response (i.e., a Gaussian function centered at the target location), context information matrix (including the HOG or deep RGB features which are extracted from the target and its background region), auxiliary variable, trainable multi-channel correlation filters, regularization parameter, and identity matrix with $N$ feature channels, respectively. Besides, the $B$ variable indicates a binary crop matrix that returns the target region as the positive sample and the real background regions as the negative ones. Finally, $M$ and $\otimes$ denote the length of the context region (vectorized image patch) and the Kronecker product, respectively.

To exploit efficient computations, the proposed method utilizes the FFT of weighted feature maps (i.e., the HOG or 3D CNN feature maps which are weighted by Gaussian window) and then constructs the context models by applying the ADMM algorithm on weighted feature map. By definition of penalty factor (with a constant $\beta$), and Lagrangian vector respectively as \cite{BACF}

\begin{equation}\label{Eq.2}
{\mathrm{\mu }}^{(i+1)}={\mathrm{min} \left({\mu }_{max},\beta {\mu }^{(i)}\right)\ }
\end{equation}
\begin{equation}\label{Eq.3}
\mathrm{\rho }\mathrm{=}\frac{1}{\sqrt{M}}\left(BF^T\otimes I_N\right)\widehat{\rho }
\end{equation}
the ADMM algorithm iteratively utilizes the two following closed-form solutions \ref{Eq.4} and \ref{Eq.5} and Lagrangian update \ref{Eq.6} as \cite{BACF}

\begin{multline}\label{Eq.4}
\hat{z}\left(t\right)=\frac{1}{\mu }\left(M\hat{y}\left(t\right)\hat{x}\left(t\right)-\hat{\rho }\left(t\right)+\mu \hat{w}\left(t\right)\right) \\ -\frac{\hat{x}\left(t\right)}{\mu \left({\hat{s}}_x\left(t\right)+M\mu \right)}\left(M\hat{y}\left(t\right)\hat{s}\left(t\right)-{\hat{s}}_{\rho }\left(t\right)+\mu {\hat{s}}_w\left(t\right)\right)
\end{multline}
\begin{equation}\label{Eq.5}
{\hat{\mathrm{w}}}^*={\left(\mu +\frac{\lambda }{\sqrt{M}}\right)}^{-1}\left(\mu z+\rho \right)
\end{equation}
\begin{equation}\label{Eq.6}
{\hat{\mathrm{\rho }}}^{(i+1)}\leftarrow {\hat{\mathrm{\rho }}}^{(i)}+\mu \left({\hat{\mathrm{z}}}^{(i+1)}-{\hat{\mathrm{w}}}^{(i+1)}\right)
\end{equation}
in which $\hat{x}$ defines the DFT of vectorized context information. Moreover, the scalers $\hat{s}$ are achieved by the multiplication of $\hat{x}\left(t\right)$ with subscript variables of them. Notice that the dimension of CNN feature maps is significantly higher than the hand-crafted feature maps that considerably reduces the convergence rate of the training process. As a result, it is noteworthy that the iteration number of the ADMM algorithm for the CNN context model must be increased in comparison with the HOG context model. It can reduce the computational efficiency of the proposed method. Hence, the proposed method prevents to construct the CNN context model until context information is not valid (described in \ref{sec:3_2}).

\subsection{Context Information Validation}
\label{sec:3_2}
So far, there is no rejection option or validation scheme to evaluate the context model in the BACF tracking method. Therefore, inaccurate estimation of context information contaminates the context model with tracking errors. 
These accumulated errors can reduce the tracking performance or lead to the drift problem. On the other hand, the persistent training of appearance model using deep RGB feature maps significantly reduces the speed of tracking methods (less than one frame per second (FPS) \cite{MDNet,CCOT}). To alleviate these issues, the proposed method extracts both the HOG and deep CNN features in each video frame. However, it does not construct a CNN context model until the HOG context model becomes inefficient or inaccurate. To detect critical situations, the proposed method enables the CNN context model exploitation when neither the rejection flag is zero, nor the HOG response map is reliable.

\begin{figure}
\centering
\includegraphics[width=8cm, height=2.5cm]{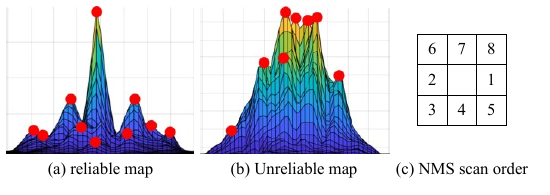}
\caption{Examples of reliable unreliable response maps which are evaluated by the 3x3-neighborhood scan order of 2D-NMS algorithm.}\label{Fig2_ReliabilityResMaps}
\vspace{-4mm}
\end{figure}

As a common problem of context-aware visual tracking methods, accumulated errors of context region information contaminate the HOG context model that aggravates the unreliable estimation of target localization. Therefore, the proposed method constructs the CNN context model when the rejection flag is enabled (i.e., set to one) to exploit the representation power of deep features in target localization. Besides, to validate the reliability of HOG response map fast (see Fig. \ref{Fig2_ReliabilityResMaps} (a, b)), the proposed method applies the fast 2D-NMS algorithm to find the considerable local maxima that might affect the accurate target localization. Due to its different scan order (see Fig. \ref{Fig2_ReliabilityResMaps} (c)) from conventional methods \cite{2DNMS}, this fast and straightforward algorithm requires at most two comparisons per pixel to detect the global ($P_g$) and local ($P_l$) maxima of the HOG response map. Then, the proposed method validates the HOG response map ($R_{HOG}$) using

\begin{equation}\label{Eq.7}
R_{HOG}=\left\{ \begin{array}{cc}
unreliable &\quad if\left\{\sum^L_{i=1}{(P^i_l\ge (T_hP_g))}\right\}>0 \\ 
reliable & otherwise \end{array}
\right.
\end{equation}
in which $L$ and $T_h$ are the number of local maxima and high threshold, respectively. According to the objective function, the response map must have a Gaussian shape with a significant peak centered on the target location when the DCF can model the target in the next frame. Hence, the HOG response map is not reliable if the number of significant peaks is more than one.

\subsection{Decision Making and Update Scheme}
\label{sec:3_3}
Despite the advantages of deep RGB feature maps, many pre-trained feature maps are often noisy or irrelevant to the visual tracking task. Hence, the proposed method utilizes a joint decision making and update scheme that prevents the contribution of irrelevant deep feature maps in the context modeling process.
The proposed method simultaneously exploits the 3D CNN feature maps (i.e., 2D spatial and 1D depth dimensions) and the FC7 feature vector which are extracted from the pre-trained VGG-19 model \cite{VGGNet}. The 3D CNN feature maps (including conv1-2, conv2-2, and conv3-4 layers) are resized and concatenated from different convolutional layers while an FC7 matrix (FC) is constructed by placing the FC7 feature vectors of valid target regions in the matrix rows as
\begin{equation} \label{Eq.8}
\mathrm{FC=}\left[ \begin{array}{c}
{FC7}^1 \\ 
\vdots  \\ 
{FC7}^V \end{array}
\right]\mathrm{=}\left[ \begin{array}{ccc}
{FC7}^1_1 & \mathrm{\cdots } & {FC7}^1_{4096} \\ 
\mathrm{\vdots } & \mathrm{\ddots } & \mathrm{\vdots } \\ 
{FC7}^V_1 & \mathrm{\cdots } & {FC7}^V_{4096} \end{array}
\right]
\end{equation}
where $V$ is the number of extracted FC7 feature vectors. To strictly focus on the best context information and effectively update the CNN context model, the proposed method exploits the FC7 feature vector of the context region. In fact, the semantic comparison of estimated context region with previous valid context regions helps the proposed method to refine itself. This semantic similarity comparison is defined as the correlation score of semantic information between the estimated context regions with the mean of prior target representations which is applied as follows.

First, the proposed method utilizes the correlation of the FC7 feature vector of estimated context region ($FC7_{estimated}$) with the mean of valid FC7 feature vectors from previous frames (which is abbreviated by FCM). In \ref{Eq.9}, $\boldsymbol{1}$ is a column vector of ones. Hence, the FC7 and rejection flags are updated via the comparison of the semantic similarity score ($FC_{score}$) with two constant thresholds ($T_h$ and $T_l$ are high and low thresholds, respectively) as follows.

\begin{equation} \label{Eq.9}
\mathrm{FCM=}{\boldsymbol{1}}^T\left(FC\right){\left({\boldsymbol{1}}^T\boldsymbol{1}\right)}^{-1}={\boldsymbol{1}}^T\left(FC\right)\left(\frac{1}{V}\right)
\end{equation}
\begin{equation} \label{Eq.10}
{\mathrm{FC}}^{update}_{score}\mathrm{=}FCM\odot {FC7}_{estimated}
\end{equation}
\begin{equation} \label{Eq.11}
{\mathrm{FC7}}^{flag}_{update}\mathrm{=}\left\{ \begin{array}{cc}
{\mathrm{FC7}}^{flag}=0 &\quad if{\mathrm{\ FC7}}^{update}_{score}<T_h \\ 
{\mathrm{FC7}}^{flag}=1 & otherwise \end{array}
\right.
\end{equation}
\begin{equation} \label{Eq.12}
{\mathrm{Rejection}}^{flag}_{update}\mathrm{=}\left\{ \begin{array}{cc}
{\mathrm{Rejection}}^{flag}=1 &\quad if{\mathrm{\ FC7}}^{update}_{score}<T_l \\ 
{\mathrm{Rejection}}^{flag}=0 & otherwise \end{array}
\right.
\end{equation}
Thus, the proposed method identifies the valid context regions when the rejection and FC7 flags are zero and one, respectively. In this case, it not only updates the FC7 matrix by adding the FC7 feature vector to its rows but also updates the 3D CNN feature maps ($\hat{x}$) using learning parameter $\eta$ as follows.

\begin{equation} \label{Eq.13}
{\hat{x}}^{\left(t\right)}=\left(1-\eta \right){\hat{x}}^{\left(t-1\right)}+\eta {\hat{x}}^{\left(t\right)}.
\end{equation}

Second, the proposed method utilizes the semantic comparison when a challenging situation has occurred, and the CNN response map is required to participate. When the fast 2D-NMS algorithm identifies more than one significant peak in the HOG response map or the rejection flag is equal to one, it detects a critical situation and enables the CNN context model. In this case, it estimates the location and scale of the context region via the construction of CNN context model (using the ADMM algorithm) and the extraction of multi-scale CNN feature maps. Then, the FC7 feature vectors of two estimated context regions (from the HOG and CNN response maps which are defined as $FC7_{candidate}$) are compared with the FCM, and the best location and scale of the target (${DM}^{l,s}_{final}$) are chosen as
\begin{equation} \label{Eq.14}
{FC}^{HOG}_{score}=FCM\odot {FC7}^1_{candidate}
\end{equation}
\begin{equation} \label{Eq.15}
{FC}^{CNN}_{score}=FCM\odot {FC7}^2_{candidate}
\end{equation}
\begin{equation} \label{Eq.16}
{DM}^{l,s}_{final}=\left\{ \begin{array}{cc}
{DM}^{l,s}_{CNN} &\quad if\ {FC}^{CNN}_{score}>{FC}^{HOG}_{score} \\ 
{DM}^{l,s}_{HOG} & otherwise \end{array}
\right.
\end{equation}
in which $\odot$, ${DM}^{l,s}_{CNN}$, and ${DM}^{l,s}_{HOG}$ are the correlation operator, estimated location and scale using the CNN and HOG context models, respectively.

\section{Experimental Results}
\label{sec:4_ExpResults}
In the proposed method, most of the implementation details of the HOG context model are the same as that of the BACF tracker. To efficiently estimate the location and scale of the target, the proposed method iteratively compares the estimation of each step of the Newton method with the previous one. It then terminates the Newton method when the difference between two successive localizations is lower than $10^{-7}$. Moreover, the high and low thresholds are experimentally set to $0.7$ and $0.4$, respectively. To appropriately train the CNN context model, the number of iterations of the ADMM algorithm is set to $20$. Furthermore, the FC7 and rejection flags are initialized to one and zero, respectively. The proposed method is implemented on the Intel I7-$6800$K CPU with $64$ GB RAM running at $3.40$ GHz. Except for the modified MatConvNet toolbox \cite{MatConvNet} that its computations were run on an NVIDIA GeForce GTX $1080$ GPU, all of the other parts of the proposed method are implemented on the CPU. Nevertheless, the proposed method has an acceptable average tracking speed of $\sim 15$ FPS (on the OTB-2015 dataset), which is considerably higher than the state-of-the-art deep learning-based visual trackers. 

The following subsections not only will provide the comprehensive evaluation of the proposed method compared with state-of-the-art methods on five large and well-known OTB-50 \cite{OTB2013}, OTB-100 \cite{OTB2015}, TC-128 \cite{TC128}, UAV-123 \cite{UAV123}, and VOT-2015 \cite{VOT-2015} visual tracking datasets but also will evaluate the ablated versions of proposed tracker to investigate the effectiveness of proposed components for visual tracking. 

\subsection{Performance Comparison: State-of-the-Art Methods}
\label{sec:4_1}
The proposed method is extensively compared with state-of-the-art visual trackers on the OTB-50, OTB-100, TC-128, UAV-123, and VOT-2015 dataset, which their benchmark results are publicly available. These visual trackers include KCF \cite{KCF}, SRDCF \cite{SRDCF}, HCFT \cite{HCFT}, HCFTs \cite{HCFTs}, LCT \cite{LCTdeep}, LCTdeep \cite{LCTdeep}, SiamFC-3s \cite{SiamFC}, DCFNet \cite{DCFNet}, DCFNet-2.0 \cite{DCFNet}, Staple-CA \cite{ContextAwareCFTracking}, SAMF-CA \cite{ContextAwareCFTracking}, BACF \cite{BACF}, DSST \cite{DSST}, Struck \cite{StruckTPAMI}, MUSTer \cite{MUSTer}, Staple \cite{Staple}, MDNet \cite{MDNet}, DeepSRDCF \cite{DeepSRDCF}, SODLT \cite{VOT-2015}, C-COT \cite{CCOT}, TADT \cite{TADT}, SA-Siam \cite{SA-Siam}, Siam-MCF \cite{Siam-MCF}, SAMF \cite{SAMF}, MEEM \cite{MEEM}, ACT \cite{ACT}, TGPR \cite{TGPR}, ASMS \cite{ASMS}, OAB \cite{VOT-2015}, SCBT \cite{SCBT}, EBT \cite{EBT}, SumShift \cite{VOT-2015}, MIL \cite{MIL}, FCT \cite{VOT-2015}, DAT \cite{DAT}, and sKCF \cite{VOT-2015}. The OTB-50, OTB-100, TC-128, UAV-123, and VOT-2015 datasets have $51$, $100$, $129$, $123$, and $60$ challenging video sequences, respectively. Also, they include challenging attributes namely illumination variation (IV), out-of-plane rotation (OPR), scale variation (SV), occlusion (OCC), partial occlusion (POC), full occlusion (FOC), deformation (DEF), motion blur (MB), fast motion (FM), in-plane-rotation (IPR), out-of-view (OV), background clutter (BC), aspect ratio change (ARC), viewpoint change (VC), camera motion (CM), similar object (SOB), and low resolution (LR). Furthermore, the video sequences of these datasets have at least $71$ and at most $3872$ frames.

The visual tracking methods are quantitatively evaluated on the OTB-50, OTB-100, TC-128, and UAV-123 datasets with two widely used tracking evaluation metrics; namely, precision and success metrics \cite{OTB2015}. While the precision metric measures the average Euclidean distance between the ground-truth and the estimated target center locations, the success metric is given by the intersection over union (IoU) of the ground-truth and the estimated bounding boxes. According to these definitions, the precision and success plots illustrate (and also rank) the percentage of video frames for which the Euclidean distance and IoU are within a given threshold (e.g., $20$ pixels and $IoU > 0.5$, respectively). Also, two well-known performance metrics of accuracy and robustness (under visual Tracking eXchange (TraX) protocol \cite{TraX}) are used to rank visual tracking methods on the VOT-2015 dataset. The accuracy metric evaluates the well-fitting of estimated bounding box compared to the ground-truth one, while the robustness metric measures the failures of methods (i.e., zero overlap of the estimated region with ground-truth one) during tracking. The VOT protocol not only re-initializes the trackers five frames after failure but also ignores ten first frames after re-initialization to reduce the bias effect on accuracy and robustness metrics. 

\begin{figure}
\centering
\includegraphics[width=11.8cm, height=4cm]{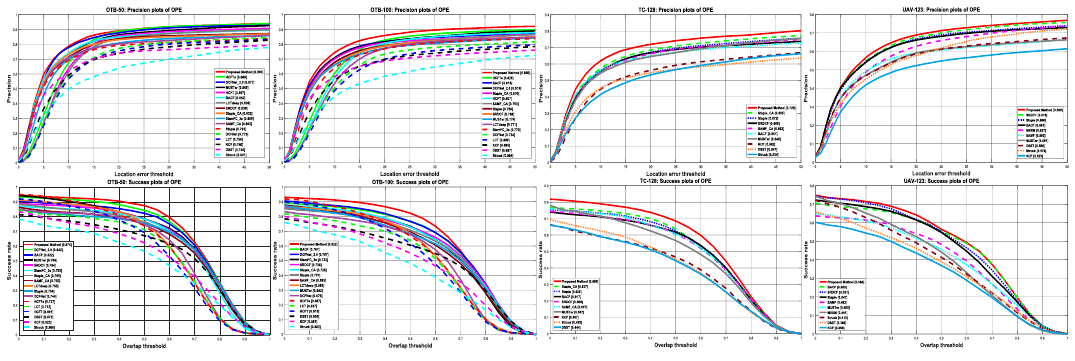}
\caption{Precision and success rate comparisons of visual tracking methods on OTB-50, OTB-100, TC-128, and UAV-123 datasets.}\label{Fig3_OverallComp}
\vspace{-4mm}
\end{figure}

\begin{figure}
\centering
\includegraphics[width=11.9cm, height=4.5cm]{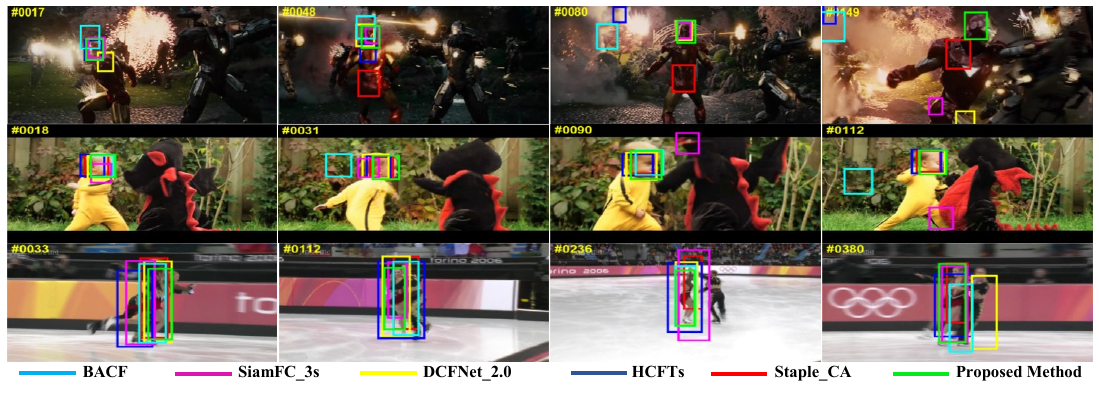}
\caption{Qualitative results of visual tracking methods on Ironman, DragonBaby, and Skating2-1 video sequences from top to bottom row, respectively.}\label{Fig4_QualComp}
\vspace{-4mm}
\end{figure}

Fig. \ref{Fig3_OverallComp} shows the precision and success plots of visual tracking methods on the OTB-50, OTB-100, TC-128 and UAV-123 datasets. These results indicate that the proposed method has improved the average precision rate up to at least $0.2\%$, and at most $24.9\%$ compared to the HCFTs and Struck, and the average success rate at least $3.2\%$, and at most $30.8\%$ compared to the DCFNet-2.0 and Struck on the OTB-50 dataset, respectively. The experimental results on the OTB-100 dataset show that the proposed method has enhanced the average precision and success rates up to at least $3.2\%$, and $5.5\%$ compared to the HCFTs, and BACF/DCFNet-2.0 and at most $27.6\%$, and $33.9\%$ compared to the Struck, respectively. Moreover, the obtained results show that the proposed method has increased the average precision rate up to at least $3.5\%$, and at most $18.6\%$ compared to the Staple-CA and Struck, the average success rate up to at least $4.9\%$ and at most $24.2\%$ compared to the Staple-CA and DSST on the TC-128 dataset, respectively. Finally, the proposed method has improved the average precision and success rates up to at least $1.3\%$ and $0.9\%$ compared to the SRDCF and BACF and at most $16.6\%$ and $19.5\%$ compared to the KCF, respectively. The attribute-based evaluations (in terms of average success rate) of visual tracking methods on OTB-100 are summarized in Table \ref{Table_AttComp}. According to this table, the proposed method has considerably achieved the best visual tracking performance at the most attributes compared to the BACF and other state-of-the-art visual tracking methods. Based on the results listed in Table \ref{Table_AttComp}, the proposed method has increased the BACF performance up to $8.1\%$, $8.5\%$, $6.7\%$, $10.9\%$, $7.8\%$, $5.7\%$, $3.4\%$, $4.4\%$, $4.6\%$, $7.5\%$, $8.3\%$ and $5.5\%$ for IV, OPR, SV, OCC, DEF, MB, FM, IPR, OV, BC, and LR attributes, respectively. Moreover, the qualitative evaluation results of the proposed method compared to the top-5 tracking methods, including Staple-CA, SiamFC-3s, HCFTs, BACF, and DCFNet-2.0, are shown in Fig. \ref{Fig4_QualComp}.

\begin{table}[!hbt]
\caption{Attribute-based evaluations on the OTB-100 dataset. [\textcolor{green}{First}, \textcolor{blue}{second}, and \textcolor{red}{third} visual tracking methods are shown in color.]} % title of Table
\centering % used for centering table
\resizebox{\textwidth}{!}{
\begin{tabular}{c c c c c c c c c c c c} 
\hline \hline 
Top-trackers & IV (38) & OPR (63) & SV (65) & OCC (49) & DEF (44) & MB (31) & FM (40) & IPR (53) & OV (14) & BC (31) & LR (10) \\ \hline \hline
Proposed Method & \cellcolor{green}0.863 & \cellcolor{green}0.793 & \cellcolor{green}0.773 & \cellcolor{green}0.804 & \cellcolor{green}0.774 & \cellcolor{green}0.809 & \cellcolor{green}0.799 & \cellcolor{blue!30}0.745 & \cellcolor{green}0.738 & \cellcolor{green}0.841 & \cellcolor{blue!30}0.720 \\ \hline 
HCFTs & 0.696 & 0.612 & 0.544 & 0.604 & 0.624 & 0.717 & 0.654 & 0.652 & 0.569 & \cellcolor{red}0.758 & 0.526 \\ \hline 
HCFT & 0.602 & 0.590 & 0.489 & 0.570 & 0.561 & 0.677 & 0.644 & 0.634 & 0.568 & 0.680 & 0.402 \\ \hline 
DCFNet-2.0 & 0.737 & \cellcolor{blue!30}0.747 & \cellcolor{blue!30}0.722 & \cellcolor{blue!30}0.747 & 0.651 & 0.734 & \cellcolor{red}0.758 & \cellcolor{green}0.747 & \cellcolor{red}0.660 & 0.738 & \cellcolor{red}0.637 \\ \hline 
SiamFC-3s & 0.716 & \cellcolor{red}0.708 & 0.697 & 0.688 & 0.638 & 0.704 & 0.719 & 0.685 & 0.627 & 0.663 & \cellcolor{green}0.750 \\ \hline 
BACF & \cellcolor{blue!30}0.782 & \cellcolor{red}0.708 & \cellcolor{red}0.706 & \cellcolor{red}0.695 & \cellcolor{blue!30}0.696 & \cellcolor{blue!30}0.752 & \cellcolor{blue!30}0.765 & \cellcolor{red}0.701 & \cellcolor{blue!30}0.692 & \cellcolor{blue!30}0.766 & \cellcolor{red}0.637 \\ \hline 
Staple-CA & 0.738 & 0.672 & 0.649 & 0.685 & \cellcolor{blue!30}0.696 & 0.718 & 0.717 & 0.676 & 0.576 & 0.724 & 0.486 \\ \hline 
Staple & 0.729 & 0.659 & 0.622 & 0.671 & \cellcolor{red}0.671 & 0.677 & 0.665 & 0.661 & 0.570 & 0.709 & 0.483 \\ \hline 
SRDCF & \cellcolor{red}0.742 & 0.663 & 0.674 & 0.680 & 0.667 & \cellcolor{red}0.747 & 0.722 & 0.657 & 0.558 & 0.701 & 0.604 \\
\hline\hline
\end{tabular}}
\label{Table_AttComp}
\end{table}

\begin{figure}
\subfigure{\includegraphics[width=5.8cm, height=4cm]{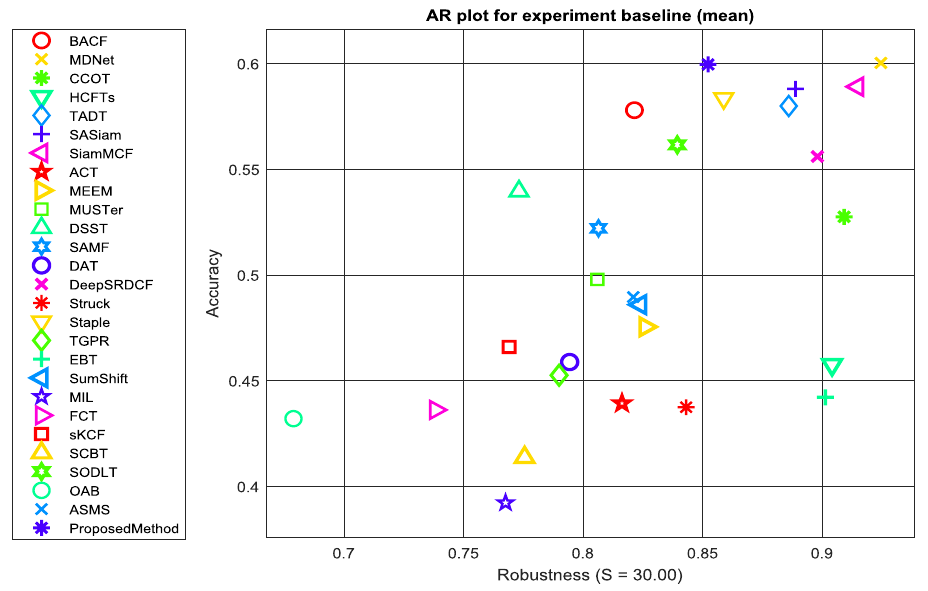}}
\subfigure{\includegraphics[width=5.8cm, height=4cm]{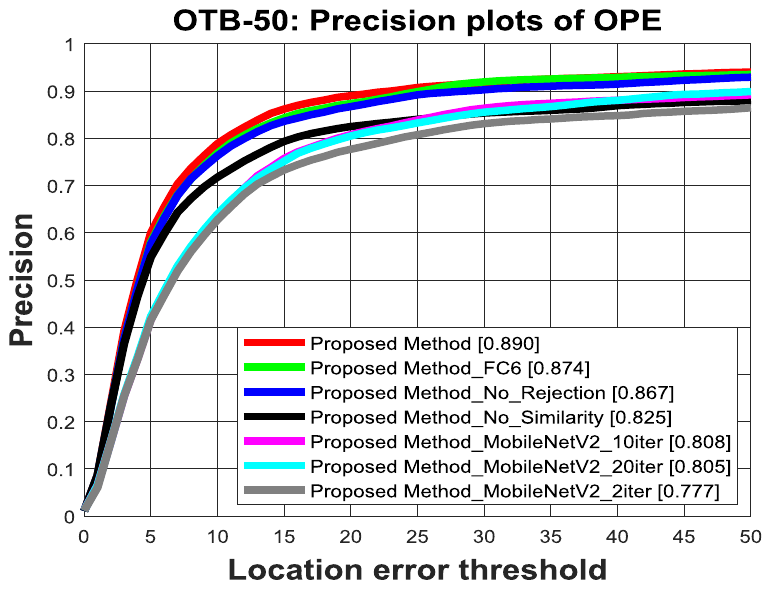}}
\caption{(left): AR ranking of visual tracking methods on VOT-2015 dataset. (right): Ablation study of proposed method on OTB-50 dataset.}\label{Fig5_VOT_Ablation}
\end{figure}

Finally, Fig. \ref{Fig5_VOT_Ablation} (left) shows the performance comparison of the proposed method compared to the state-of-the-art visual tracking methods in terms of accuracy and robustness rank (AR rank) on the VOT-2015 dataset. The proposed method has enhanced the BACF performance up to $2.7\%$ and $7.2\%$ in terms of accuracy and robustness metrics, respectively. According to these results, the proposed method not only has considerably improved the tracking performance of the BACF but also has almost achieved the accuracy of the MDNet (which has one FPS speed). Furthermore, the proposed method has significantly improved the robustness of the BACF. The robustness result of proposed method originates from the moderate robustness of BACF and transferred deep features faced with some challenging attributes such as LR, MB, and CM. The proposed method has outperformed the other context-aware visual trackers (e.g., \cite{ContextAwareCFTracking,BACF}) which have a competitive performance against deep learning-based tracking methods. Although the context information has reduced the unwanted boundary effects of correlation filters to achieve a higher tracking performance, it may not handle realistic scenarios in which several challenging attributes occur, simultaneously. However, the proposed method tries to detect these situations and robustly estimate the context region. It also represents better results compared to the LCT, LCTdeep, and MUSter trackers that exploit short- and long-term memory storages. Furthermore, the proposed method has provided better quantitative and qualitative results compared to the recent deep feature-based visual trackers that even exploit an end-to-end framework (e.g., SiamFC-3s, DCFNet, and DCFNet-2.0).

The desirable performance of the proposed tracking method is due to the following reasons. First, it simultaneously uses the CNN feature maps to achieve a better target representation. The CNN context model is not exploited until the HOG context model cannot accurately estimate the target location. This adaptive exploitation leads to a remarkable performance along with an acceptable tracking speed. Second, the proposed method learns two context models, separately. Although the estimated target regions are validated to update the rejection and FC7 flags, the HOG model is updated in all video frames. On the other hand, the update process of CNN feature maps (which are involved in the construction of the CNN model) is dependent on the status of flags. In fact, the HOG context model represents a more flexible target model, whereas the CNN context model constructs a strict valid target model. This duality helps the proposed method to avoid the accumulated tracking errors (i.e., the drift problem). Third, the proposed method can enjoy the rejection option when the target is in extremely challenging situations (such as occlusion or out-of-view). This option prevents the update process of the CNN context model and allows the proposed method to re-detect the target (without any detection module) if it is in the search area. Finally, it exploits the semantic information along with low-level feature maps which are beneficial to capture intra-class variations. These semantic comparisons can modify the estimated location of the target when the HOG context model tends to drift towards the background.

The desirable performance of the proposed tracking method is due to the following reasons. First, it simultaneously uses the CNN feature maps to achieve a better target representation. The CNN context model is not exploited until the HOG context model cannot accurately estimate the target location. This adaptive exploitation leads to a remarkable performance along with an acceptable tracking speed. Second, the proposed method learns two context models, separately. Although the estimated target regions are validated to update the rejection and FC7 flags, the HOG model is updated in all video frames. On the other hand, the update process of CNN feature maps (which are involved in the construction of the CNN model) is dependent on the status of flags. In fact, the HOG context model represents a more flexible target model, whereas the CNN context model constructs a strict valid target model. This duality helps the proposed method to avoid the accumulated tracking errors (i.e., the drift problem). Third, the proposed method can enjoy the rejection option when the target is in extremely challenging situations (such as occlusion or out-of-view). This option prevents the update process of the CNN context model and allows the proposed method to re-detect the target (without any detection module) if it is in the search area. Finally, it exploits the semantic information along with low-level feature maps which are beneficial to capture intra-class variations. These semantic comparisons can modify the estimated location of the target when the HOG context model tends to drift towards the background.

\subsection{Performance Comparison: Ablation Study}
\label{sec:4_2}
To investigate the effectiveness of proposed components, four ablated trackers including the proposed method: i) without rejection flag, ii) without semantic similarity comparison, iii) which exploits FC6 feature vector (instead of FC7), iv) which exploits the first convolutional layer (shallow features) of lightweight MobileNetV2 (instead of HOG features) with different ADMM iteration numbers are evaluated on the OTB-50 dataset (see Fig. \ref{Fig5_VOT_Ablation} (right)). According to these results, replacement the HOG features with shallow features extracted from the MobileNetV2 has the most degradation on the proposed method. As previously mentioned in \cite{MDNet}, the pre-trained CNNs have not been trained for video processing applications, and these networks may provide some deficiency when they exploit for transfer learning. Moreover, the incorrect number of ADMM iterations may cause under-fitting or overfitting of target modeling. Following, no exploitation of semantic similarity comparison, no utilization of rejection option, and replacement the FC7 with FC6 feature vector have the most destructive effects on the proposed method, respectively.

\section{Conclusion}
\label{sec:5_Conclusion}
An adaptive context-aware visual tracking method which exploits both hand-crafted and deep RGB feature maps for context modeling was proposed. With the aid of the fast 2D non-maximum suppression algorithm and deep semantic information, the proposed method detects challenging situations through video frames. When a challenging situation is detected, the proposed method constructs a CNN context model (which is according to the valid deep RGB feature maps) to increase the accuracy of the target localization by using the HOG and CNN context models. Otherwise, the HOG context model was used to increase visual tracking speed. Based on the proposed context information validation, an update scheme allowed the proposed method to either update the CNN feature maps or avoid the contaminated CNN feature maps. The extensive experimental results on the OTB-50, OTB-100, TC-128, UAV-123, and VOT-2015 visual tracking datasets demonstrated that the proposed method not only improves the visual tracking performance, but it also enjoys acceptable tracking speed compared to the state-of-the-art tracking methods.\\

\noindent\textbf{Acknowledgement:} This work was partly supported by a grant (No. $96013046$) from Iran National Science Foundation (INSF).\\

\noindent\textbf{Compliance with Ethical Standards:} All authors declare that they have no conflict of interest.

\vspace{-4mm}
\bibliographystyle{plainnat}
\bibliography{ref.bib}
\end{document}